# A HYBRID COA/ε-CONSTRAINT METHOD FOR SOLVING MULTI-OBJECTIVE PROBLEMS

Mahdi parvizi, Elham Shadkam and Niloofar jahani

Department of Industrial Engineering, Faculty of Eng.; Khayyam University, Mashhad, Iran

## ABSTRACT

*In this paper, a hybrid method for solving multi-objective problem has been provided. The proposed method is combining the ε-Constraint and the Cuckoo algorithm. First the multi objective problem transfers into a single-objective problem using ε-Constraint, then the Cuckoo optimization algorithm will optimize the problem in each task. At last the optimized Pareto frontier will be drawn. The advantage of this method is the high accuracy and the dispersion of its Pareto frontier. In order to testing the efficiency of the suggested method, a lot of test problems have been solved using this method. Comparing the results of this method with the results of other similar methods shows that the Cuckoo algorithm is more suitable for solving the multi-objective problems.*

## 1. INTRODUCTION

In the single-objective optimization it is assumed that the decision makers connect to a single purpose Such as maximizing the profit, minimizing the costs, minimizing the waste, maximizing the market share etc. But in the real world, the decision maker checks more than a single objective. For example in order to study the production level in a company, if only the profit would be examined and all other objectives such as customer satisfaction, staff satisfaction, the production diversity, market share etc would be rejected, the results won't be reliable. So using the multi-objective decision making (MODM) is necessary. Finding an optimized answer that covers all of the restrictions together is impossible in multi-objective problems. So using the Pareto frontier, reliable answers for a multi-objective problem will be obtained. There are many different ways for solving multi-objective problems. These ways divide in two groups. Combined methods (all of the objectives acts as a single one) and the limited methods (one of the objective function will be kept and other ones would be act as the restriction).

Ehrgott and Gandibleux studied on the approximate and the accurate problems related to the combination method of multi-objective problems [1]. Hannan and Klein submitted an algorithm for solving multi-objective integer linear programming. This algorithm use to eliminate the extra known dominant solutions [2]. Leumanns et al. submitted a meta-heuristic algorithm in order to find approximate effective solutions of multi-objective integer programming, using the ε-Constraint [3]. Sylva and Crema submitted a solution for finding the set of non-dominant vectors in multi-objective integer linear programming [4]. Arakaw et al. combined the GDEA and the GA methods to generate the efficient frontier in multi-objective optimization problems. [5] Deb used the evolutionary algorithms for solving the multi-objective algorithms [6]. Nakayama drew the Pareto frontier of the multi-objective optimization algorithms using DEA (Data Envelopment Analysis) [7]. Agarwal drew the Pareto frontier of the multi-objective optimization algorithms using GA (Genetic Algorithm) [8]. Vincova used the DEA in order to find the Pareto frontier [9]. Reyes-Sierra investigated the solution of multi-objective optimization algorithm using the particle swarm algorithm [10]. Seiford and Tone helped the multi-objective optimization algorithm using DEA and publishing related software [11]. Pham solved the multi-objective optimization algorithm using the Bee Algorithm [12]. Durillo and Garc'ıa-Nieto investigated a new solution for multi-objective optimization algorithm based on the particle swarm algorithm [13]. Yun studied the solution of multi-objective optimization algorithm using the GA and DEA. Also he found the Pareto frontiers of efficient points using this method [14]. Yang used the Cuckoo optimization algorithm in order to find the Pareto frontiers [15]. Gorjestani et al. proposed a COA multi objective algorithm using DEA method [16].

This article submits a hybrid algorithm that uses the advantages of both the Cuckoo algorithm and the ε-Constraint method simultaneously. The submitted algorithm solves the multi-objective problems for allowable εs using the Cuckoo algorithm and the Matlab software. At last for each iteration, it finds a Pareto answer and linking these answers draw the Pareto frontier. This method draws a better Pareto frontier than other similar methods. In the second section, the Cuckoo algorithm will be introduced. The third section explains the multi-objective algorithm and the ε-Constraint method. In the fourth section, the suggested hybrid algorithm of this article will be investigated in details. Test problems and their solution with similar algorithms compares in the fifth section then in the last section, the conclusion and the future offers will be submitted.

## 2. THE CUCKOO ALGORITHM INTRODUCTION

The cuckoo search was expanded by Xin-She Yang and Suash Deb in 2009. After that the Cuckoo optimization algorithm was submitted by Ramin Rajabioun in 2011 [17]. This algorithm applied in several researches such as production planning problem [18](Akbarzadeh and Shadkam, 2015), portfolio selection problem [19](Shadkam *et al.*, 2015), evaluation of organization efficiency [20](Shadkam and Bijari, 2015), evaluation of COA [21](Shadkam and Bijari, 2014) and so on

Flowchart of the Cuckoo algorithm is given in the figure 1

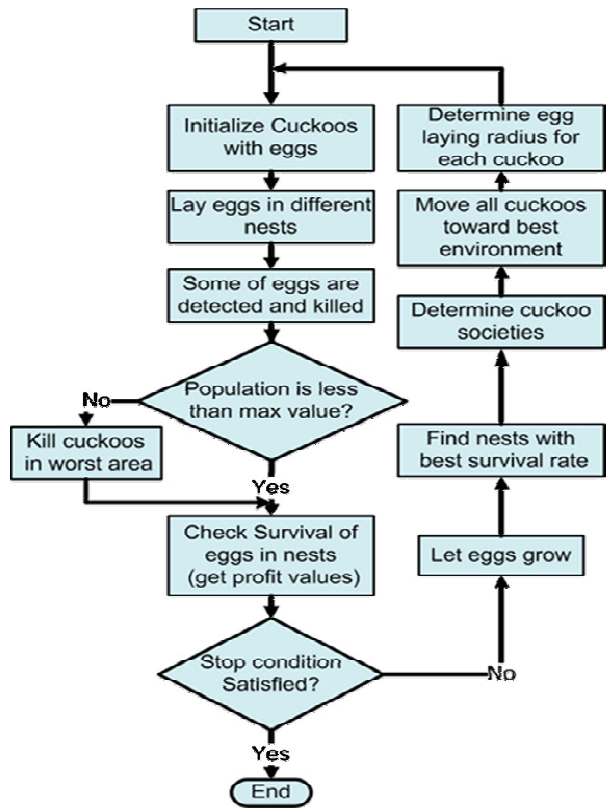

Figure1: the Cuckoo algorithm flowchart

For more information refer to [17].

## 3. THE MULTI-OBJECTIVE ALGORITHM AND THE ε-CONSTRAINT METHOD:

General form of a multi-objective optimization problem is as (1):

$Max\ (Min) = f_1(x_j)$
$Max\ (Min) = f_2(x_j)$
⋮
$Max(Min) = f_k(x_j)$ (1)
s.t.
$g_i(x_j) <=> b_i, \quad i = 1, 2 \ldots m$
$x_j \geq 0, \quad j = 1, 2, \ldots, n$

In the multi-objective problems, we face some objectives in contrast of single-objective algorithms that has just one objective. In this model, k is the number of objective functions that can be either max or min and m is the number of restrictions and n is the number problem's variables.

In the multi-objective optimization problems there is not a certain answer that optimizes all of the objective functions simultaneously. For this reason, the Pareto optimal concept is introduced.

The Pareto optimal concept explanation is that $\bar{x}_* = (x_1 \ldots, x_2, x_n)$ is an optimal pareto. If for each allowable $\bar{x}$ and i={1,2,..k}, we have (for minimizing problem): $\forall_{i \epsilon I}(f_i(\bar{x}_*) \leq f_i(\bar{x}_i))$

Then $\bar{x}_*$ will be the optimal Pareto that n is the number of decision making variables and k is the number of objective functions. In other words, $\bar{x}_*$ is an optimal Pareto if there is no other $\bar{x}$ vector that doesn't make at least one objective function worse in order to improve some of the objective functions.

## 4. ε-CONSTRAINT METHOD

In this method, one of the different objective functions will be selected and other objective functions will act as the restrictions considering a specific constraint and the problem changes into a single-objective problem. Using different εs results optimal pareto answers.

General form of this method is given as (2).

*Min F(X)={$f_i(x), \ldots, f_n(x)$}*
*s.t.*
*$g(x) <=> b$*
*$x \geq 0$*

(2)

*Min F(X)=$f_i(x)$*
*s.t.*
*$g(x) <=> b$*
*$f_j(x) \leq \varepsilon_j, j \neq i, j = 1,..,n$*
*$x \geq 0$*

If the objective function is max, the constraint is $f_j(x) \geq \varepsilon_j$. Selecting the ε is the most important thing in this method because the answers are so sensitive to this parameter. So the selected ε must be in range of $f_j^{min} \leq \varepsilon_j \leq f_j^{max}$ for each objective function.

## 5. THE COA/ ε-CONSTRAINT HYBRID ALGORITHM

Step 1: first according to the objective function of main problem, the mathematical model will be written based on the ε-Constraint method and the problem is converted from multi-objective to single-objective problem.
Step 2: the obtained function from the ε-Constraint method will be described as the meta-heuristic Cuckoo algorithm function.
Step 3: the iterations including the εs for solving the main problem is formed for the Cuckoo algorithm, and this loop will be iterate until the Cuckoo algorithm ends.
Step 4: according to the first laying time and the initial number of cuckoos, a matrix will be formed from the habitats in the beginning of implementing the Cuckoo algorithm.
Step 5: the obtained function from step 2 gets the habitats matrix as its input data and finds the objective problem magnitude according to the new restrictions for each habitat.
Step 6: the Cuckoo algorithm sorts the habitats according to their quantity and objective functions as usual and the rest of the tasks will be the same as it is described in references number [17].
Step 7: in each iteration of the loop, the habitat that earns the most quantity of the objective function called best cuckoo and will be saved in a different matrix.
Step 8: after exiting the formed loop, the functions $f_1$ and $f_2$ calculate for each saved points.

Step 9: a plot of the $f_1$ and $f_2$ functions quantity will be drawn that is the Pareto frontier of the main multi-objective optimization problem.

## 6. SOLVING TEST PROBLEMS

A number of test functions have been provided that can help to validate the suggested approach in table 4.

Table 4. Test problems

| | *Objective function* | *constraints* |
|---|---|---|
| 1 | $min\ f_1 = x_1$ <br> $min\ f_2 = x_2$ | $4 \leq (x_2 - 2)^2 + (x_1 - 2)^2$ <br> $x_2, x_1 \geq 0$ |
| 2 | $min\ f_1 = 2x_1 - x_2$ <br> $min\ f_2 = -x_1$ | $(x_1 - 1)^3 + x_2 \leq 0$ <br> $x_2, x_1 \geq 0$ |
| 3 | $min\ f_1 = x_1$ <br> $min\ f_2 = x_2$ | $-x_2 - 3x_1 x_1^3 \geq 0$ <br> $x_1 \geq -1,\ x_2 \leq 2$ |
| 4 | $min\ f_1 = 4x_1 + 4x_2$ <br> $min\ f_2 = (x_1 - 5)^2 + (x_2 - 5)^2$ | $(x_2)^2 + (x_1 - 5)^2 \geq 25$ <br> $x_2, x_1 \geq 0$ |
| 5 | $min\ f_1 = (\sum_{i=1}^{2} -10 \exp(-0.2\sqrt{x_i^2 + x_{i+1}^2})$ <br> $min\ f_2 = \sum_{i=1}^{3}[|x_i|^{0.8} + 5 \sin(x_i^3)]$ | $-5 \leq x_i \leq 5,\ i=1,2,3$ |
| 6 | $min\ f_1 = x_1$ <br> $min\ f_2 = x_2$ | $\cos(16 \arctan(\frac{x_1}{x_2})) \geq 0 - 1 - 0.1 x_2^2 + x_1^2$ <br> $(x_2 - 0.5)^2 + (x_1 - 0.5)^2 \geq -0.5$ <br> $x_1 \geq 0,\ x_2 \geq \pi$ |
| 7 | $min\ f_1 = 1 - \exp(-\sum_{i=1}^{n}(x_i - \frac{1}{\sqrt{n}})^2)$ <br> $min\ f_2 = 1 - \exp(-\sum_{i=1}^{n}(x_i + \frac{1}{\sqrt{n}})^2)$ | $-4 \leq x_i \leq 4,\ i = 1,2$ |
| 8 | $min\ f_1 = x_1$ <br> $min\ f_2 = \frac{(1 + x_2)}{x_1}$ | $9x_1 + x_2 \geq 6$ <br> $9x_1 + x_2 \geq 1$ <br> $x_1 \in [0.1, 1], x_2 \in [0, 5]$ |
| 9 | $min\ f_1 = (x_1 - 2)^2 + (x_2 - 1)^2 + 2$ <br> $(x_2 - 1)^2 - min\ f_2 = 9x_1$ | $3x_2 - x_1 \geq -10$ <br> $(x_2)^2 + (x_1)^2 \geq 225$ <br> $x_1 \in [-20, 20], x_2 \in [-20, 20]$ |

According to high importance of the input parameters of meta-heuristic algorithm and its effect on the final answer, the parameters of the Cuckoo algorithm for solving any problems are given below:

Number of initial population=5, minimum number of eggs for each=2, maximum number of eggs for each cuckoo= 4, number of clusters=1.

**Test problem 1:** [22,17]

For solving the example using the ε-Constraint method, one of the objective functions will be kept and the other one will be added to the constraints as it mentioned before. For this test function, we keep $f_1$ in the objective function and add $f_2$ to the constraint and the problem will be as the equation (3):

$$min\ f_1 = x_1$$
s.t.
$$(x_2 - 2)^2 + (x_1 - 2)^2 \geq 4 \quad (3)$$
$$x_2 \geq \varepsilon$$
$$x_2 x_1 \geq 0$$

In order to find the allowable range of ε, $f_2$ will be solved once with min function and once with max function. The allowable range will be the $0 \leq \varepsilon \leq 4$.

For finding allowable εs with the pace of 0.01, the problem will be solved using the Cuckoo algorithm and the Matlab software for 400 iterations. The Pareto frontier is shown in figure 2. Also the results of finding the Pareto frontier using the similar methods are shown in this figure too.

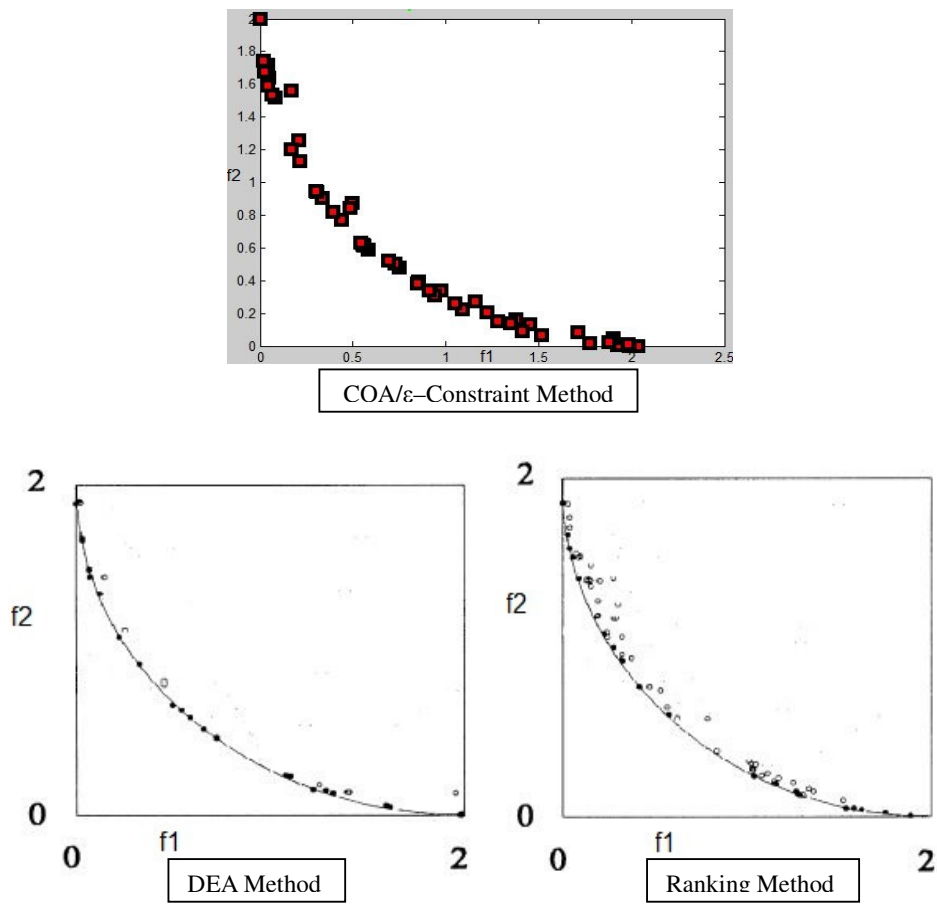

Figure 2. Comparing the suggested method with other methods

32

**Test problem 2:** [22,17]

The converted problem is as the equation (4)

$$min\ f_1 = 2x_1 - x_2$$
s.t.
$$x_2 + (x_1 - 1)^3 \leq 0$$
$$-x_1 \geq \varepsilon \qquad\qquad (4)$$
$$x_2 x_1 \geq 0$$

The allowable range of ε will be the $-1 \leq \varepsilon \leq 0$ and the pace is 0.0025. The Pareto frontier after 400 iterations is shown in figure 3.

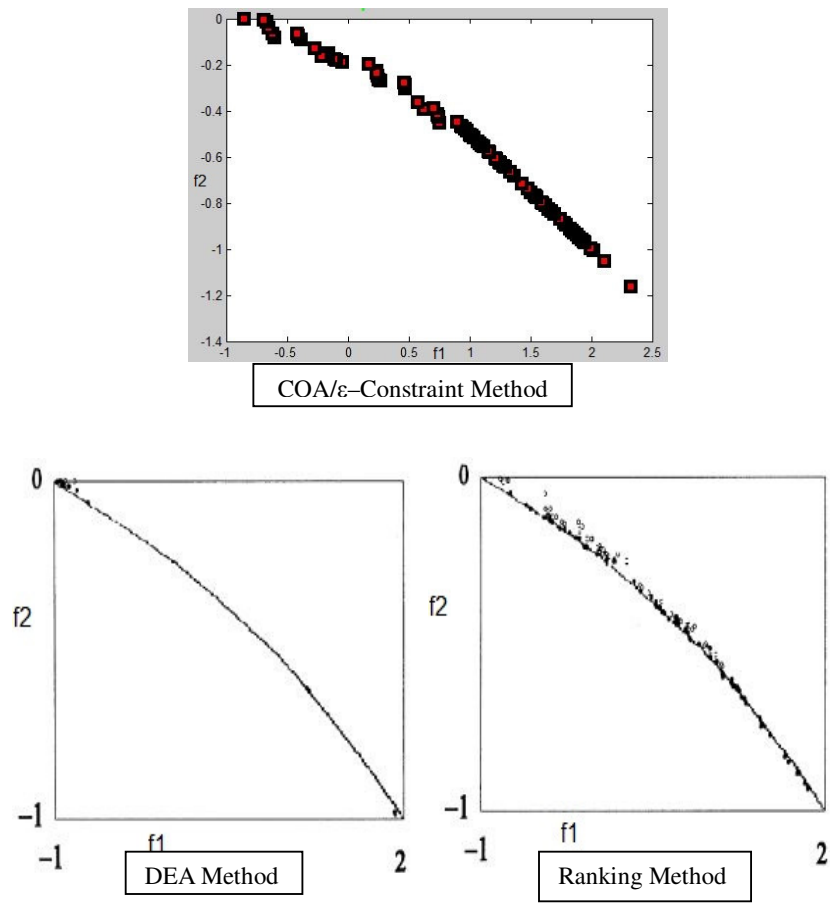

Figure 3. Comparing the suggested method with other methods

**Test problem 3:** [22,17]

The allowable range of ε will be the $-2 \leq \varepsilon \leq 2$ and the pace is 0.01 and the Pareto frontier after 400 iterations is shown in figure 4.

International Journal in Foundations of Computer Science & Technology (IJFCST) Vol.5, No.5, September 2015

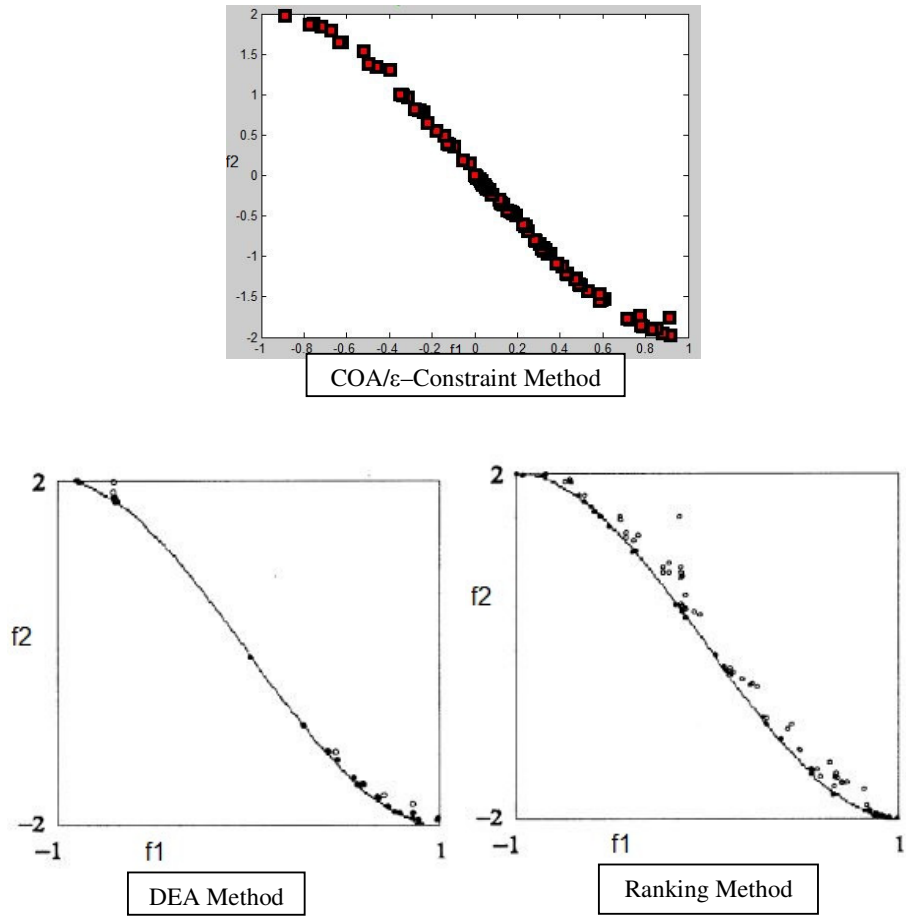

Figure 4. Comparing the suggested method with other methods

**Test problem 4**:[23,18]

The allowable range of ε will be the $0 \leq \varepsilon \leq 50$ and the pace is 0. 125. The Pareto frontier after 400 iterations is shown in figure 5.

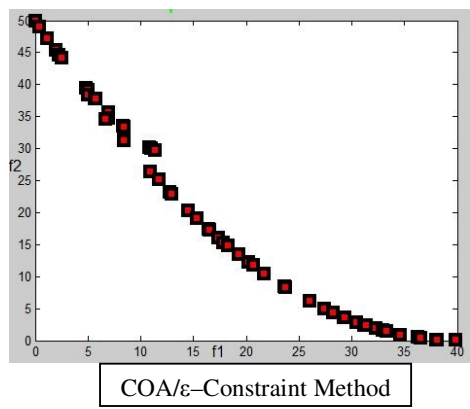

COA/ε–Constraint Method

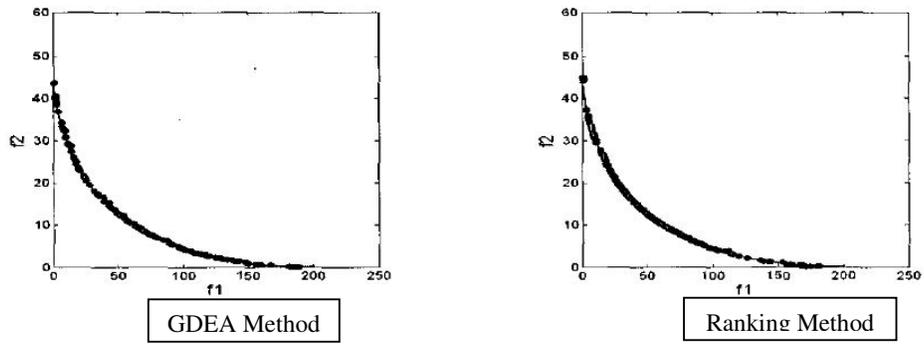

Figure 5. Comparing the suggested method with other methods

**Test problem 5**:[24,19]

The allowable range of ε will be the $-11 \leq ε \leq 20$ and the pace is 0.0775. The Pareto frontier after 400 iterations is shown in figure 6.

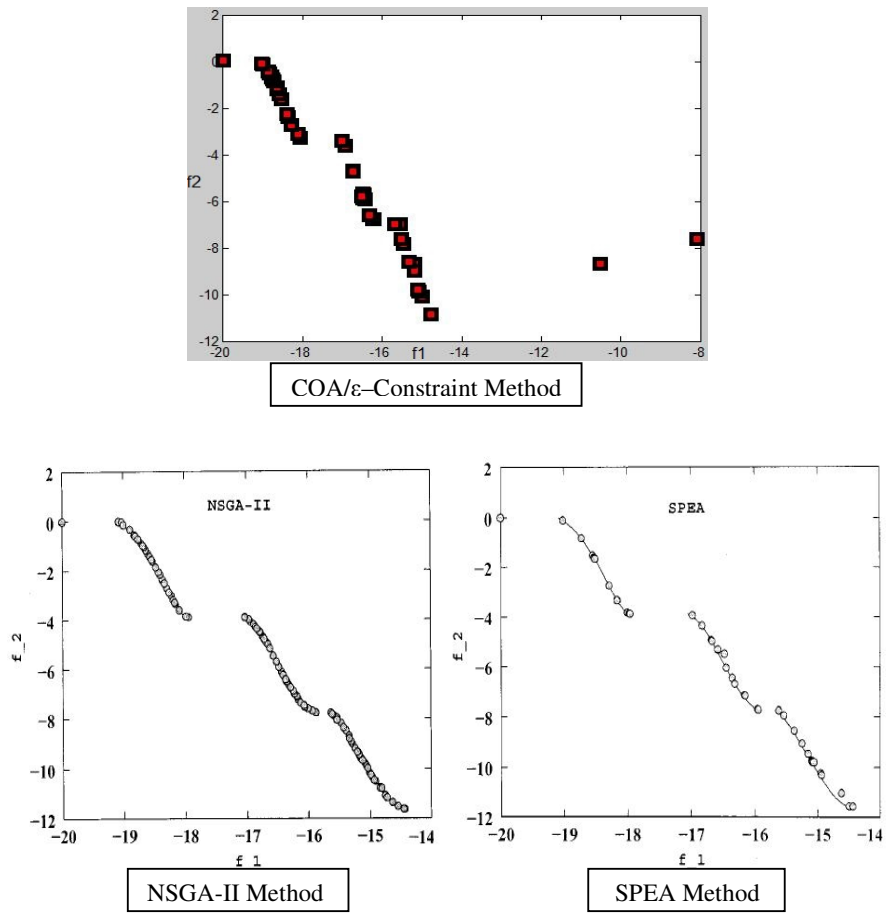

Figure 6. Comparing the suggested method with other methods

35

**Test problem 6**:[24,19]

The allowable range of ε will be the 0 ≤ε≤ 1.2 and the pace is 0. 008. The Pareto frontier after 400 iterations is shown in figure 7.

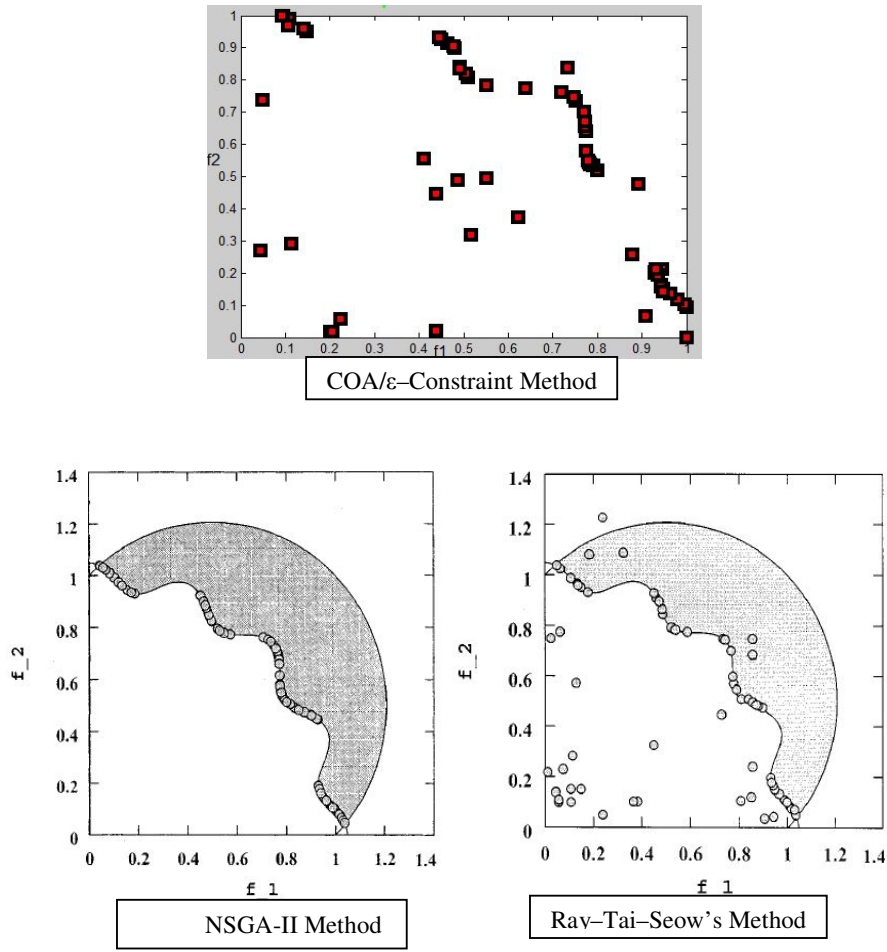

Figure 7. Comparing the suggested method with other methods

**Test problem 7:** [24,19]

The allowable range of ε will be the −25 ≤ε≤ 1 and the pace is 0. 065. The Pareto frontier after 400 iterations is shown in figure 8.

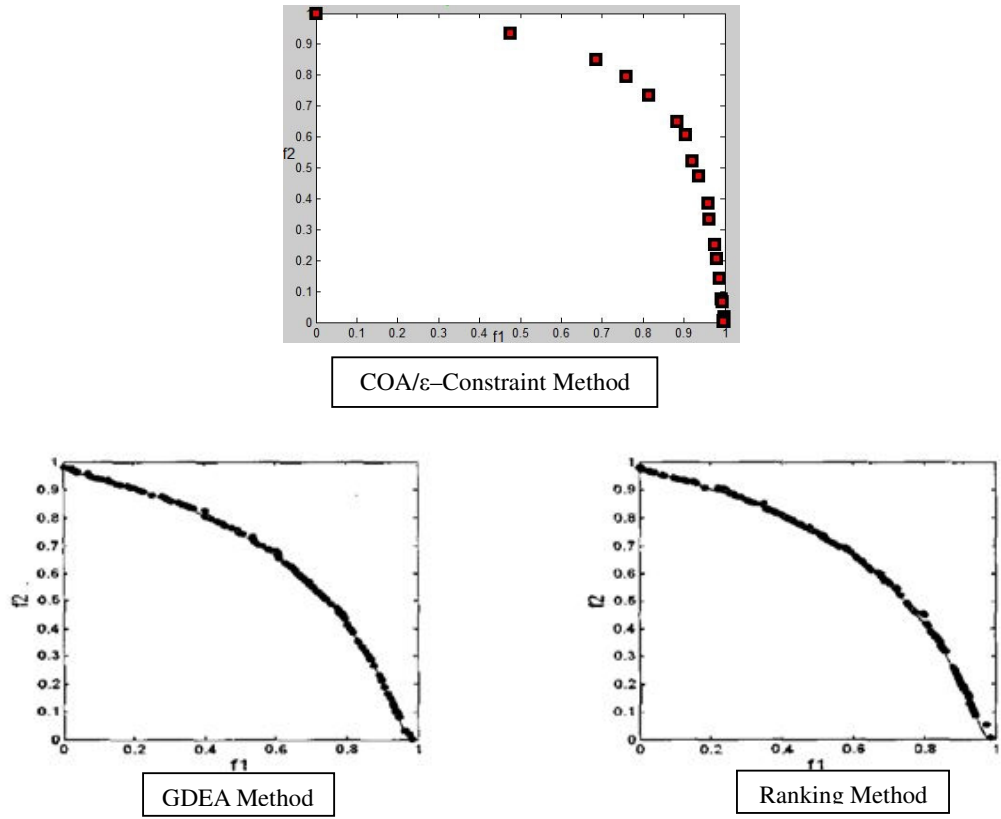

Figure 8. Comparing the suggested method with other methods

**Test problem 8**: [24,19]

The allowable range of ε will be the $1 \leq \varepsilon \leq 9$ and the pace is 0.02. The Pareto frontier after 400 iterations is shown in figure 9.

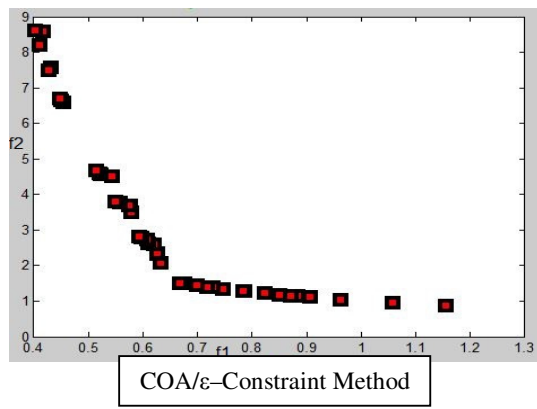

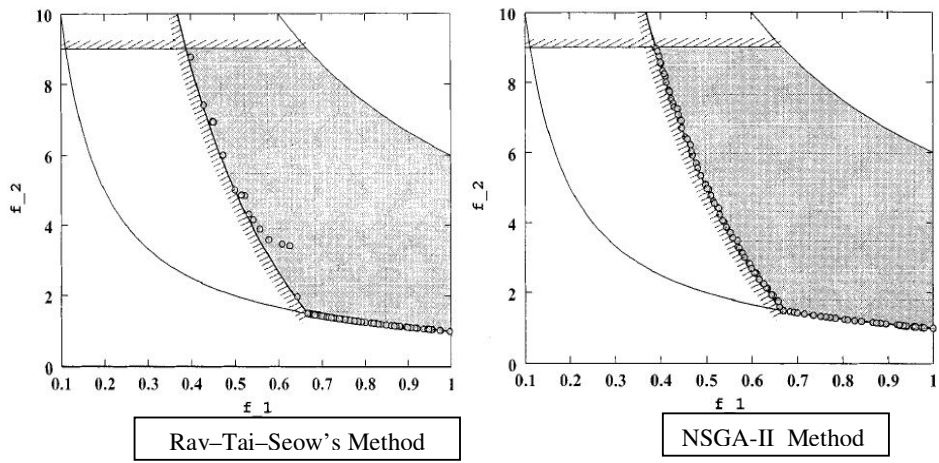

| Rav–Tai–Seow's Method | NSGA-II Method |

Figure 9. Comparing the suggested method with other methods

**Test problem 9:**[24,19]

The allowable range of ε will be the $-196 \leq \varepsilon \leq 72$ and the pace is 2.68. The Pareto frontier after 400 iterations is shown in figure 10.

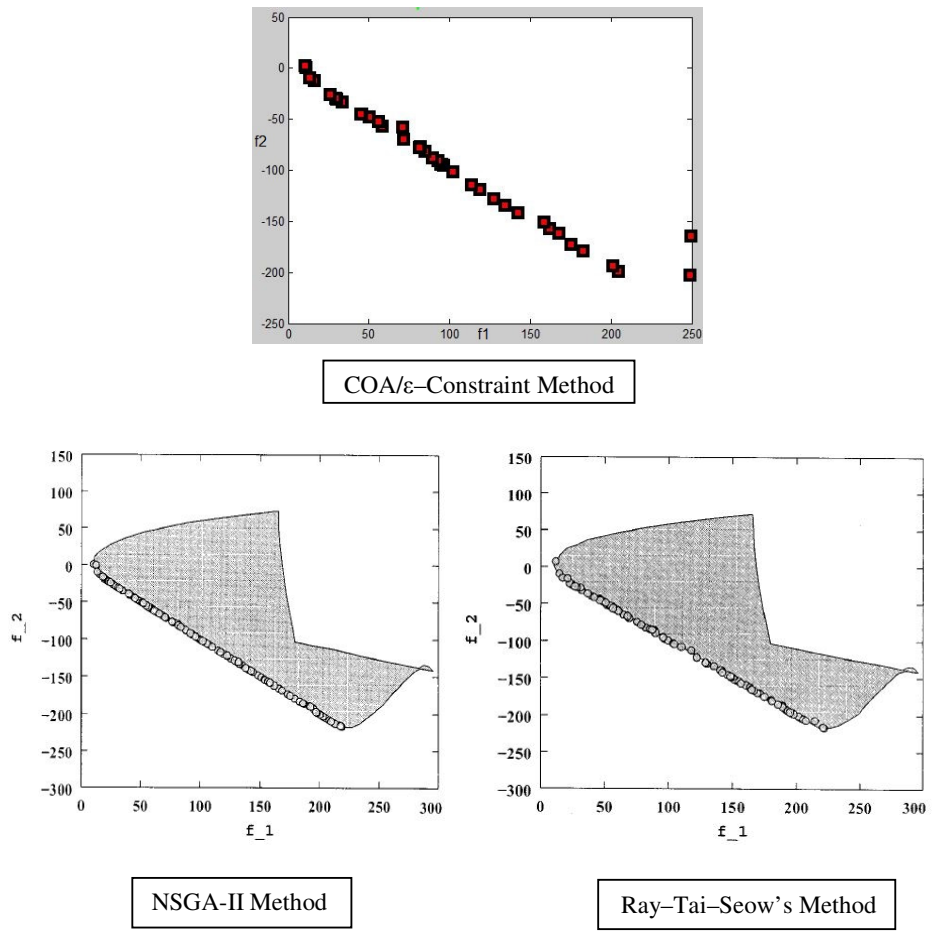

COA/ε–Constraint Method

| NSGA-II Method | Ray–Tai–Seow's Method |

Figure 10. Comparing the suggested method with other methods

According to the Pareto frontier resulted from different functions, it is evident that the suggested method provides uniform and exact frontiers in fewer iterations than other similar methods.

## 7. CONCLUSION

In this paper, we presented a hybrid method for solving multi-objective problems using the Cuckoo algorithm and the ε-Constraint method. According to the obtained results from the proposed method and comparing the obtained Pareto frontiers with the results of similar methods such as GDEA/GA, DEA/GA, RANKING, NSGA-II, Ray–Tai–Seow's and SPEA, we concluded that not only the Cuckoo algorithm finds better Pareto frontiers but also, it needs shortest time to give the Pareto frontier. Pareto frontier of proposed method has more dispersion than the other similar algorithms. So the COA/ε-Constraint method is a suitable and reliable method for solving multi-objective optimization problems. In the future, solving the problems with more objectives, multi-objective allocation problem and multi-objective problems of project controlling with minimizing the time and cost target would be in order.